\documentclass{article}

\usepackage[preprint]{neurips_2026}

\usepackage[utf8]{inputenc}
\usepackage[T1]{fontenc}
\usepackage{hyperref}
\usepackage{url}
\usepackage{booktabs}
\usepackage{amsfonts}
\usepackage{amsmath}
\usepackage{amssymb}
\usepackage{nicefrac}
\usepackage{microtype}
\usepackage{xcolor}
\usepackage{graphicx}
\usepackage{subcaption}
\usepackage{multirow}
\usepackage{adjustbox}

\title{Ask Early, Ask Late, Ask Right: When Does Clarification Timing Matter for Long-Horizon Agents?}

\author{%
  Anmol Gulati\thanks{Corresponding authors: \texttt{anmol.b.gulati@pwc.com}, \texttt{elias.lumer@pwc.com}} \mdseries\qquad
  Hariom Gupta \qquad
  Elias Lumer \\
  Sahil Sen \qquad
  Vamse Kumar Subbiah \\[0.5em]
  PricewaterhouseCoopers U.S.
}

\begin{document}

\maketitle

\begin{abstract}
Long-horizon AI agents execute complex workflows spanning hundreds of sequential actions, yet a single wrong assumption early on can cascade into irreversible errors. When instructions are incomplete, the agent must decide not only \emph{whether} to ask for clarification but \emph{when}, and no prior work measures how clarification value changes over the course of execution. We introduce a forced-injection framework that provides ground-truth clarifications at controlled points in the agent's trajectory across four information dimensions (goal, input, constraint, context), three agent benchmarks, and four frontier models (three per benchmark; one on a single benchmark only; 84 task variants; 6,000+ runs). Counter to the common intuition that ``earlier is always better,'' we find that the value of clarification depends sharply on \emph{what} information is missing: goal clarification loses nearly all value after 10\% of execution (pass@3 drops from 0.78 to baseline), while input clarification retains value through roughly 50\%. Deferring \emph{any} clarification type past mid-trajectory degrades performance below never asking at all. Cross-model Kendall $\tau$ correlations (0.78--0.87 among models sharing identical task coverage; 0.34--0.67 across the full 4-model panel) confirm these timing profiles are substantially task-intrinsic. A complementary study of 300 unscripted sessions reveals that no current frontier model asks within the empirically optimal window, with strategies ranging from over-asking (52\% of sessions) to never asking at all. These empirical demand curves provide the quantitative foundation that existing theoretical frameworks require but have lacked, and establish concrete design targets for timing-aware clarification policies. Code and data will be publicly released.
\end{abstract}

\section{Introduction}

Long-horizon AI agents now execute complex multi-step workflows spanning code repair, enterprise automation, and tool orchestration, often completing tens to hundreds of sequential actions with minimal human oversight~\citep{yao2023react, xu2024theagentcompany, jimenez2024swebench, bandi2026mcpatlas, lumer2025toolagentsurvey}. When user instructions are incomplete, these agents face a fundamental decision: continue executing under uncertainty, or pause to request clarification. Consider an enterprise agent tasked with generating a quarterly report that spends 30 actions building the wrong table before realizing the user meant fiscal quarters, not calendar quarters. A single clarifying question at step two could have prevented the wasted work entirely. The cost of this decision scales with trajectory length, as each additional action taken under wrong assumptions compounds the risk of irreversible errors and wasted computation~\citep{lumer2026don}.

Recent benchmarks show that underspecification consistently degrades agent performance and that clarification channels enable partial recovery~\citep{pu2026lhaw, elfeki2026hilbench}, yet existing work treats clarification as a binary capability: the agent either can or cannot ask. Neither LHAW~\citep{pu2026lhaw} nor HiL-Bench~\citep{elfeki2026hilbench} addresses the temporal dimension, though LHAW's failure mode analysis independently identifies timing as a problem, finding that agents ask at inappropriate points in the trajectory. Classical value-of-information theory~\citep{howard1966information} and HCI interruption research~\citep{adamczyk2004ifnotnow, mark2008cost} both predict that clarification benefit should vary with timing, yet no empirical measurement of this interaction exists for modern LLM agents. Asking too early risks querying about information that is inferable from context not yet observed. Asking too late means the agent has already committed irreversible actions based on wrong assumptions.

\begin{figure}[t]
  \centering
  \includegraphics[width=\linewidth]{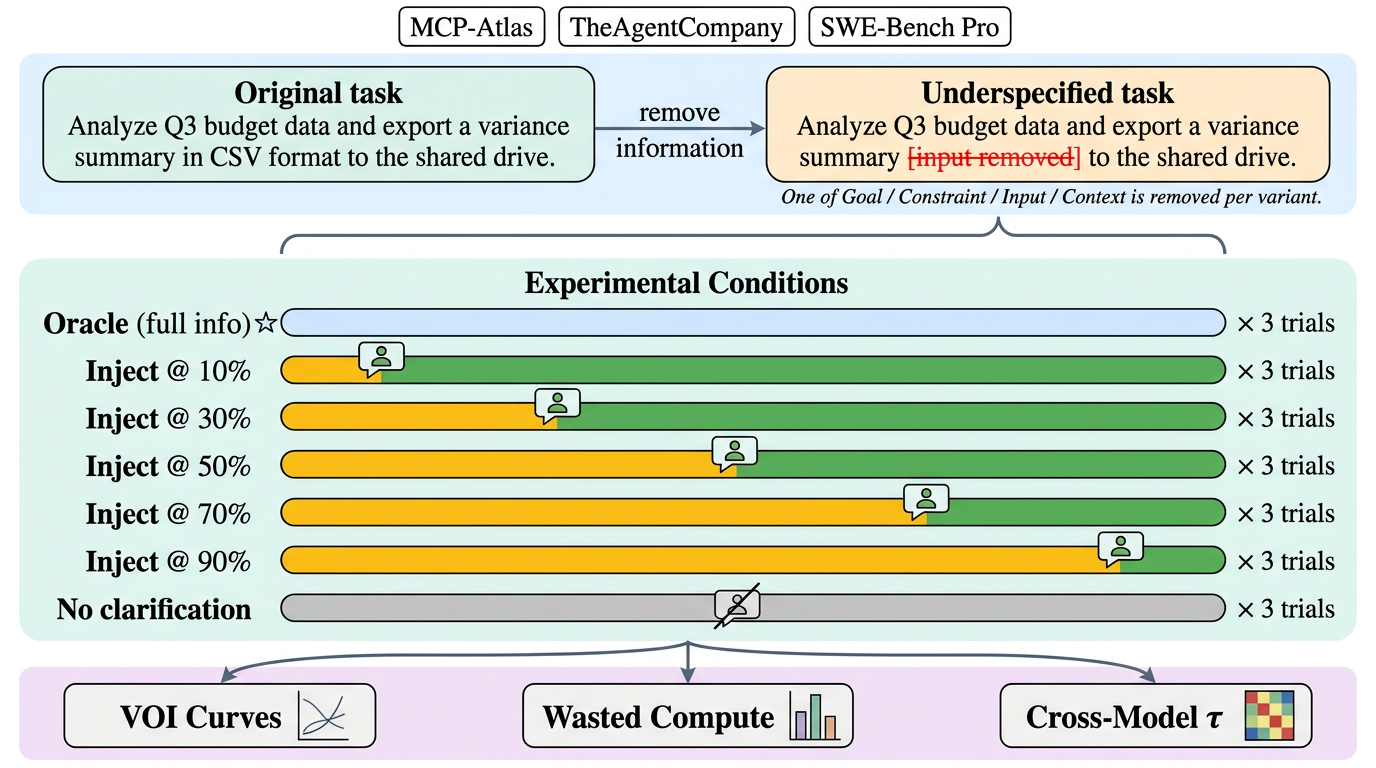}
  \caption{Overview of the forced-injection experimental framework. We inject ground-truth clarifications at controlled points along an oracle-calibrated action budget, measuring task success (pass@3) at each injection timing across four information dimensions.}
  \label{fig:overview}
\end{figure}

We isolate the pure timing effect with a forced-injection framework that provides ground-truth clarifications at controlled trajectory points, independent of the agent's ability to detect ambiguity, across 84 underspecified task variants, three benchmarks, and four frontier models (Figure~\ref{fig:overview}). Our experiments yield two findings that challenge common assumptions about clarification timing. First, the value of clarification is not simply ``earlier is better''; it depends sharply on what type of information is missing, with goal clarification losing nearly all value after 10\% of execution while input clarification retains value through roughly 50\%. Second, despite this structure, no current frontier model naturally asks within the empirically optimal window for any dimension, with strategies ranging from over-asking (52\% of sessions) to never asking at all.

Our contributions are:
\begin{enumerate}
    \item We establish that clarification timing interacts strongly with information dimension: goal information has a narrow early window (${\leq}$10\% of execution), input information is recoverable through ${\sim}$50\%, and constraint benefit depends on whether an oracle gap exists, with cross-model consistency (Kendall $\tau$ of 0.78--0.87 among models sharing identical task coverage; 0.34--0.67 on the broader panel) confirming these profiles are substantially task-intrinsic. Wasted compute grows steadily with delay, providing a complementary cost measure (Section~\ref{sec:results}).
    \item We quantify the gap between current model asking behavior and these empirically optimal windows: across 300 unscripted sessions, natural ask timings are severely late for goal tasks, and one frontier model never asks at all, establishing concrete design targets for timing-aware clarification policies.
\end{enumerate}

\section{Related Work}

Prior work addresses whether clarification helps, when information is theoretically valuable, and how agents should decide to ask. We identify a common blind spot: none varies clarification \emph{timing} as an independent variable.

\subsection{Clarification in Agent Benchmarks}

A growing body of work establishes that underspecification degrades agent performance and that clarification channels enable partial recovery. LHAW~\citep{pu2026lhaw} constructs 285 underspecified task variants across four information dimensions and shows that models exhibit widely divergent clarification strategies. HiL-Bench~\citep{elfeki2026hilbench} complements this by penalizing both over-asking and missed escalation. Shorter-horizon benchmarks study related phenomena: QuestBench~\citep{li2025questbench}, AmbigQA~\citep{min2020ambigqa}, ClariQ~\citep{aliannejadi2020clariq}, UserBench~\citep{qian2025userbench}, and $\tau$-bench~\citep{yao2024taubench}. Tamkin et al.~\citep{tamkin2023task} show that task ambiguity systematically degrades model behavior, and Kadavath et al.~\citep{kadavath2022know} establish that LLMs have limited self-assessment of uncertainty. Several approaches aim to close this gap: confidence-based selective clarification~\citep{kuhn2023clam}, uncertainty-aware planning~\citep{hu2024uncertainty}, proactive pre-execution questioning~\citep{zhang2024askbeforeplan}, and simulation-based ask-or-answer training~\citep{wu2025collabllm}. In reinforcement learning, ask-for-help policies~\citep{clouse1996integrating} and DAgger~\citep{ross2011dagger} query experts under uncertainty, Plaut et al.~\citep{plaut2025catastrophe} prove that agents must query at a linear rate or risk catastrophe, and recent LLM systems learn \emph{whether} to ask via counterfactual training~\citep{liu2025cart}, process reward models~\citep{min2025selfregulation}, or interactive feedback~\citep{pan2025benchmarks}. All optimize a confidence threshold but assume information is equally useful at any trajectory point. Broader long-horizon agent benchmarks~\citep{zhou2024webarena, trivedi2024appworld, liu2024agentbench, xie2024osworld, mialon2023gaia} similarly do not vary clarification timing.

\subsection{Theoretical Foundations Predict Timing Should Matter}

The decision to seek clarification mid-trajectory is a special case of the value-of-information (VOI) problem~\citep{lindley1956information, howard1966information}: should the agent pay a cost (interruption, latency) to reduce uncertainty before acting? Russell and Wefald~\citep{russell1991right} formalize this as metareasoning, and the active-learning paradigm~\citep{settles2009active} studies when to query an oracle to maximize learning efficiency. In HCI, Adamczyk and Bailey~\citep{adamczyk2004ifnotnow} demonstrate that task outcomes depend on \emph{when} an interruption occurs, and Mark et al.~\citep{mark2008cost} quantify the recovery cost of poorly timed interruptions. Recent work operationalizes these ideas for LLM agents: Rao and Daum\'{e}~\citep{rao2018evpi} apply expected value of perfect information to rank clarification questions, Dong et al.~\citep{dong2026voi} propose a decision-theoretic framework, Kobalczyk et al.~\citep{kobalczyk2025active} cast clarification as Bayesian experimental design, Suri et al.~\citep{suri2025structured} use structured uncertainty for tool-calling agents, Lumer et al.~\citep{lumer2025memtool} optimize short-term memory management across multi-turn tool interactions, and PRISM~\citep{fu2026prism} learns a cost-sensitive gate for proactive interventions. These frameworks all predict that clarification value should depend on timing, yet none provides the empirical measurements our forced-injection experiments produce.

\section{Hypotheses: Trajectory Commitment}
\label{sec:model}

We develop a simple model predicting \emph{why} clarification timing should interact with information dimension. A task is specified by a parameter vector $\theta = (\theta_{\text{goal}}, \theta_{\text{input}}, \theta_{\text{con}}, \theta_{\text{ctx}})$, one component per dimension. The agent observes an underspecified version $\tilde{\theta}$ in which one component $\theta_d$ is missing, and executes a trajectory $\tau = (a_1, \ldots, a_T)$. At injection time $t$, the agent receives the true value of $\theta_d$. We define the \emph{commitment} $C_d(t) \in [0,1]$ as the fraction of actions $a_1, \ldots, a_t$ that causally depend on dimension $d$. The value of clarification at time $t$ is bounded by the recoverable portion:
\begin{equation}
  \text{VOI}_d(t) \;\leq\; \text{VOI}_d(0) \cdot \bigl(1 - C_d(t)\bigr),
  \label{eq:voi_bound}
\end{equation}
since committed actions cannot be undone. The shape of $C_d(t)$ thus upper-bounds the timing decay profile for each dimension; the actual VOI may decay faster if reconciliation costs exceed the recoverable portion. We argue that $C_d(t)$ grows at different rates due to the causal structure of typical workflows: \emph{Goal} and \emph{context} condition all subsequent actions: early approach selection cascades through the entire trajectory, predicting concave, front-loaded commitment. \emph{Input} affects only data-dependent steps; agents can partially infer inputs through exploration, predicting approximately linear commitment. \emph{Constraints} impose rules that may be invisible until violated; actions taken without constraint knowledge may actively conflict with it, so late injection can be \emph{disruptive}, with reconciliation cost exceeding the information's value.
\begin{table}[t]
  \caption{Dataset and experimental setup. Action budgets are oracle-calibrated per model and variant. Total runs include all conditions and trials across available models.}
  \label{tab:dataset}
  \centering
  \begin{tabular}{lcccc}
    \toprule
    Benchmark & Variants & Dimensions & Action Budget & Total Runs \\
    \midrule
    MCP-Atlas & 36 & goal, constraint, input, context & 6--20 & 3,024 \\
    TheAgentCompany & 30 & goal, constraint, input & 6--49 & 1,890 \\
    SWE-Bench Pro & 18 & goal, constraint, input, context & 1--121 & 1,134 \\
    \bottomrule
  \end{tabular}
\end{table}

\begin{table}[t]
  \caption{Frontier models evaluated. API identifiers are provided for reproducibility. Clarification archetypes are derived from the natural-ask protocol (Section~\ref{sec:natural_ask}). A fourth model, DeepSeek V3.2, is evaluated on MCP-Atlas only (see Section~\ref{sec:benchmarks}).}
  \label{tab:models}
  \centering
  \small
  \begin{tabular}{llcc}
    \toprule
    Model & API Identifier & Provider & Archetype \\
    \midrule
    GPT-5.2 & \texttt{gpt-5.2-2025-12-11} & OpenAI & Over-clarifier (52\%) \\
    Claude Sonnet 4.5 & \texttt{claude-sonnet-4-5} & Anthropic & Balanced (23\%) \\
    Gemini 3 Flash & \texttt{gemini-3-flash-preview} & Google & Under-clarifier (0\%) \\
    DeepSeek V3.2 & \texttt{deepseek-v3-2} & DeepSeek & N/A (MCP-Atlas only) \\
    \bottomrule
  \end{tabular}
\end{table}

These predictions yield two falsifiable hypotheses. \textbf{H1} (Dimension-dependent timing): Goal and context clarification are front-loaded (steep VOI decay), while input clarification decays gradually. \textbf{H2} (Constraint attenuation): On benchmarks where an oracle gap exists, constraint clarification yields VOI that declines with delay, with reconciliation costs offsetting some but not all of the information's value. We test both in Section~\ref{sec:results}.

\section{Experimental Design}

\subsection{Overview and Hypotheses}

We test the hypotheses derived from the trajectory commitment model (Section~\ref{sec:model}): \textbf{H1} predicts that goal and context clarification are front-loaded while input clarification decays gradually, and \textbf{H2} predicts that constraint clarification yields VOI that declines with delay on benchmarks where an oracle gap exists, with reconciliation costs offsetting some but not all of the information's value. We additionally test whether, for each dimension, there exists a \emph{point of no return} beyond which clarification no longer improves over the no-clarification baseline.

For each of 84 underspecified task variants (full breakdown in Appendix~\ref{sec:dataset_details}), we run seven conditions with three trials per condition across available models (four on MCP-Atlas, three on TheAgentCompany and SWE-Bench Pro), yielding over 6,000 total experiment runs across all conditions, models, and trials (Appendix~\ref{sec:compute_cost}). The seven conditions are: (1) \emph{oracle}, where the agent receives the fully-specified original prompt (upper bound); (2) \emph{no-clarification}, where the agent receives the underspecified prompt and must complete the task without additional information (lower bound); and (3--7) five \emph{injection} conditions at 10\%, 30\%, 50\%, 70\%, and 90\% of the trajectory, where the missing information is provided mid-execution.

\begin{figure}[t]
  \centering
  \includegraphics[width=\linewidth]{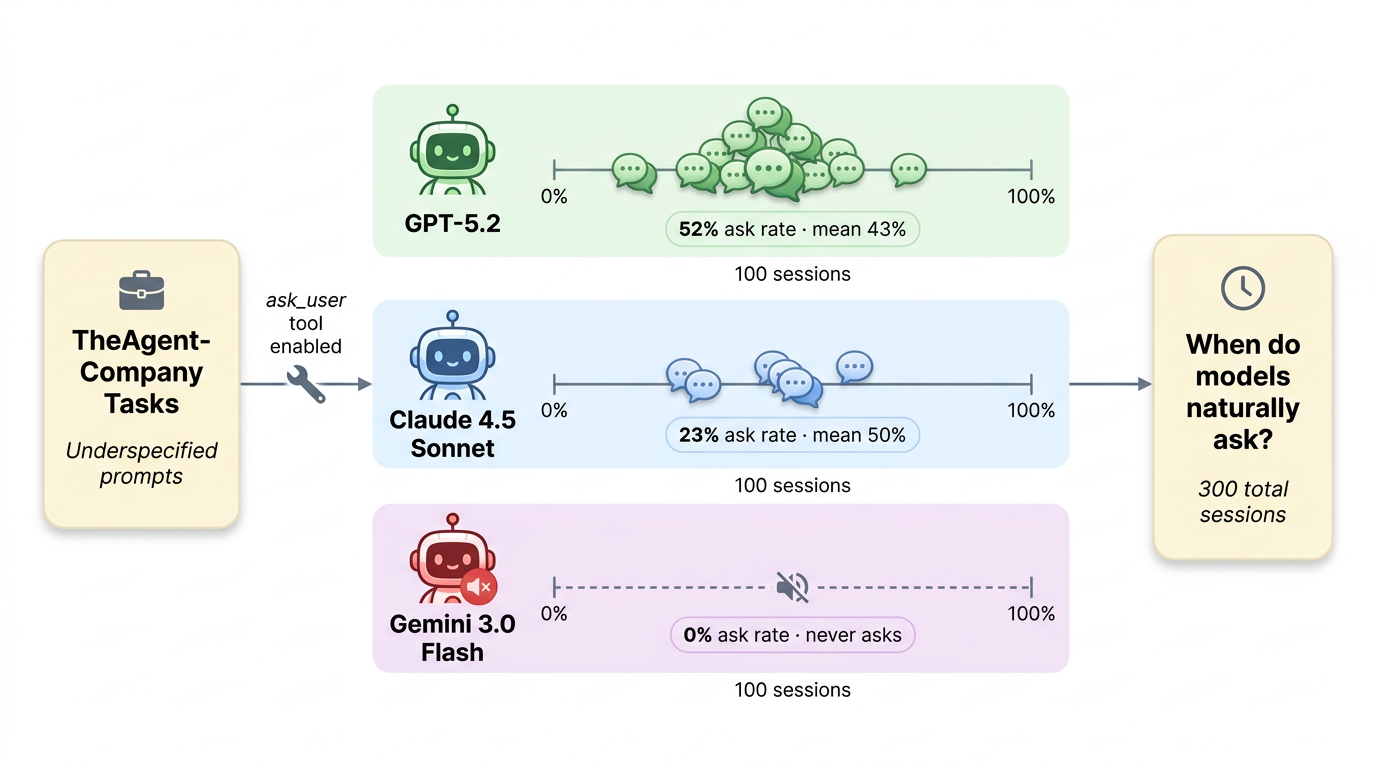}
  \caption{Natural ask protocol. Three models each run 100 sessions on underspecified TheAgentCompany tasks with the \texttt{ask\_user} tool enabled. We report two distinct percentages: the \emph{ask rate} (fraction of sessions in which the model asked at least once) and the \emph{mean ask timing} (how far through the trajectory the first ask occurred, as a percentage of total actions). GPT-5.2 asks frequently (ask rate 52\%, mean timing 43\%), Claude Sonnet 4.5 asks selectively (ask rate 23\%, mean timing 50\%), and Gemini 3 Flash never asks (ask rate 0\%).}
  \label{fig:natural_ask_protocol}
\end{figure}

\subsection{Benchmarks and Models}
\label{sec:benchmarks}

We draw our 84 variants from a stratified subset of LHAW's~\citep{pu2026lhaw} underspecified tasks, spanning three benchmarks that cover distinct agent domains (Table~\ref{tab:dataset}):

\textbf{MCP-Atlas}~\citep{bandi2026mcpatlas, lumer2025scalemcp, lumer2025tool} (36 variants, 4 models) evaluates tool-use competency across real MCP servers with short trajectories (6--20 actions). All four information dimensions are represented. This is the only benchmark evaluating all four models including DeepSeek V3.2, as its standard tool-calling APIs require no benchmark-specific harness.

\textbf{TheAgentCompany}~\citep{xu2024theagentcompany} (30 variants, 3 models) tests enterprise workflows on a simulated corporate platform with longer trajectories (6--49 actions) involving web, database, and tool interactions. Three dimensions are covered (goal, constraint, input). DeepSeek V3.2 is excluded due to harness integration limitations.

\textbf{SWE-Bench Pro}~\citep{scaleai2025swebenchpro} (18 variants, 3 models) evaluates code-repair tasks on real GitHub issues (1--121 actions). DeepSeek V3.2 is excluded; the remaining three models each have both injection and baseline data, yielding $n{=}12$--31 (variant, model) units per injection cell depending on dimension (Appendix~\ref{sec:swe_limitations}).

LHAW defines four information dimensions: \emph{goal} (unclear deliverable), \emph{constraint} (missing rules or thresholds), \emph{input} (unspecified data source), and \emph{context} (absent domain knowledge). Variants are stratified by dimension and ambiguity class: \emph{outcome-critical} (missing information causes failure), \emph{divergent} (valid but unintended output), and \emph{benign} (agent succeeds despite the gap).

\subsection{Forced Injection Protocol}

Because tasks range from 6 to 121 actions, we calibrate injection points per model and task. For model $m$ and variant $v$, let $B_{m,v}$ denote the mean oracle trajectory length (rounded to the nearest integer). The injection action for timing $t \in \{0.1, 0.3, 0.5, 0.7, 0.9\}$ is $a_{\text{inject}} = \max(1, \lfloor B_{m,v} \cdot t \rfloor)$. For SWE-Bench Pro tasks where $B_{m,v} = 1$, all five injection conditions collapse to action~1; these are retained to preserve sample size but contribute zero timing variance.

At action $a_{\text{inject}}$, a synthetic user message containing the ground-truth \texttt{removed\_segments} is inserted at the next clean turn boundary (Appendix~\ref{sec:injection_details},~\ref{sec:example_trajectory}). The \texttt{ask\_user} tool is disabled, so clarification arrives only through the injection. Messages read naturally (e.g., ``By the way, I should have mentioned: the target format is CSV, not JSON.''). Each action corresponds to one tool call or one agent message turn.

\subsection{Natural Ask Protocol}
\label{sec:natural_ask}

To assess whether models naturally ask within the optimal timing window, we run a complementary 300-session study on TheAgentCompany tasks with the \texttt{ask\_user} MCP tool enabled. Each of three models (Claude Sonnet 4.5, GPT-5.2, Gemini 3 Flash) completes 100 sessions across seven base tasks (a stratified subset of the 30 TAC variants, selected to cover goal-dimension and input-dimension variants in roughly equal proportion across all three ambiguity classes). When invoked, a simulated user responds with the ground-truth missing information matching the forced-injection content. We record the action step of each \texttt{ask\_user} call as a percentage of total trajectory length (Figure~\ref{fig:natural_ask_protocol}).

\begin{figure}[t]
  \centering
  \includegraphics[width=0.9\linewidth]{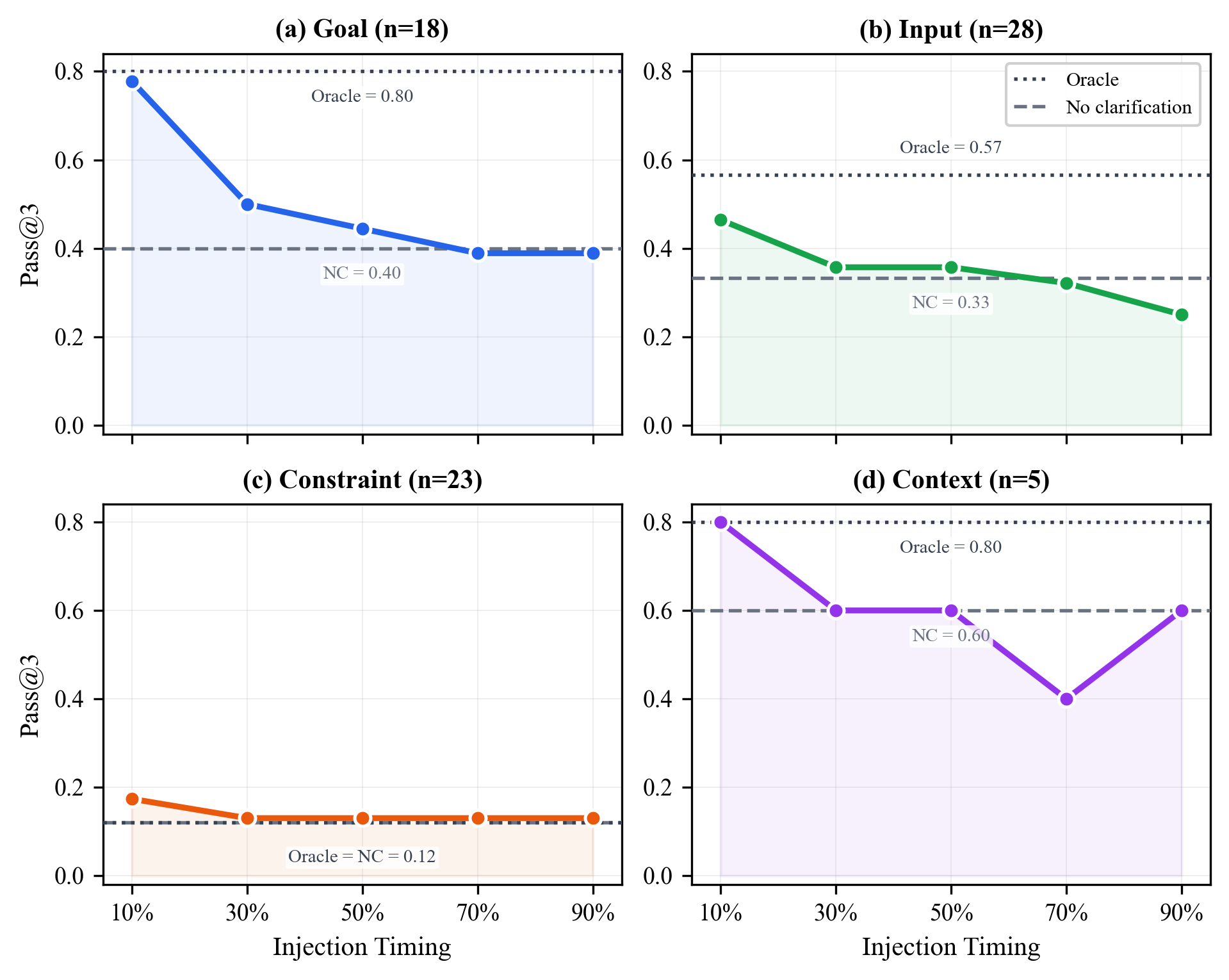}
  \caption{MCP-Atlas VOI curves by information dimension ($n{=}5$--28 per dimension; see subplot titles). Goal clarification is dramatically front-loaded (0.78 at 10\% vs.\ 0.40 NC), input shows gradual decline, and constraint shows minimal benefit over baseline. Dashed: no-clarification baseline; dotted: oracle. Per-benchmark curves and aggregate curves appear in Appendix~\ref{sec:per_benchmark_voi} and~\ref{sec:aggregate_interpret}.}
  \label{fig:voi_mcp}
\end{figure}

\subsection{Metrics}

We evaluate with four complementary metrics:

\textbf{Pass@3.} Following the HumanEval formulation~\citep{chen2021evaluating}, we compute the probability that at least one of $k=3$ trials succeeds for each (variant, model, condition) cell. We report the mean pass@3 across units within each experimental group.

\textbf{Wasted compute.} For injection conditions, we count pre-injection actions whose effects are absent from the oracle (fully-specified) trace, normalized by total pre-injection actions. This quantifies the fraction of early work rendered useless by late clarification.

Actions absent from the oracle trace may still be indirectly useful (e.g., exploratory schema queries that inform the agent's understanding). We therefore treat this metric as a conservative upper bound on waste, appropriate for quantifying the cost ceiling of late clarification rather than claiming every non-oracle action is worthless. Qualitative analysis (Appendix~\ref{sec:qualitative}) confirms that some ``wasted'' exploration partially compensates for missing specifications, consistent with input's gradual VOI decline.

\textbf{Point of no return.} The latest injection timing at which pass@3 significantly exceeds the no-clarification baseline (one-sided permutation test, $p < 0.05$), stratified by dimension and ambiguity class.

\textbf{Cross-model consistency.} Kendall's $\tau$ rank correlation~\citep{kendall1938measure} of per-variant pass@3 vectors across model pairs, measuring whether timing effects are task-intrinsic or model-dependent.

\begin{figure}[t]
  \centering
  \includegraphics[width=\linewidth]{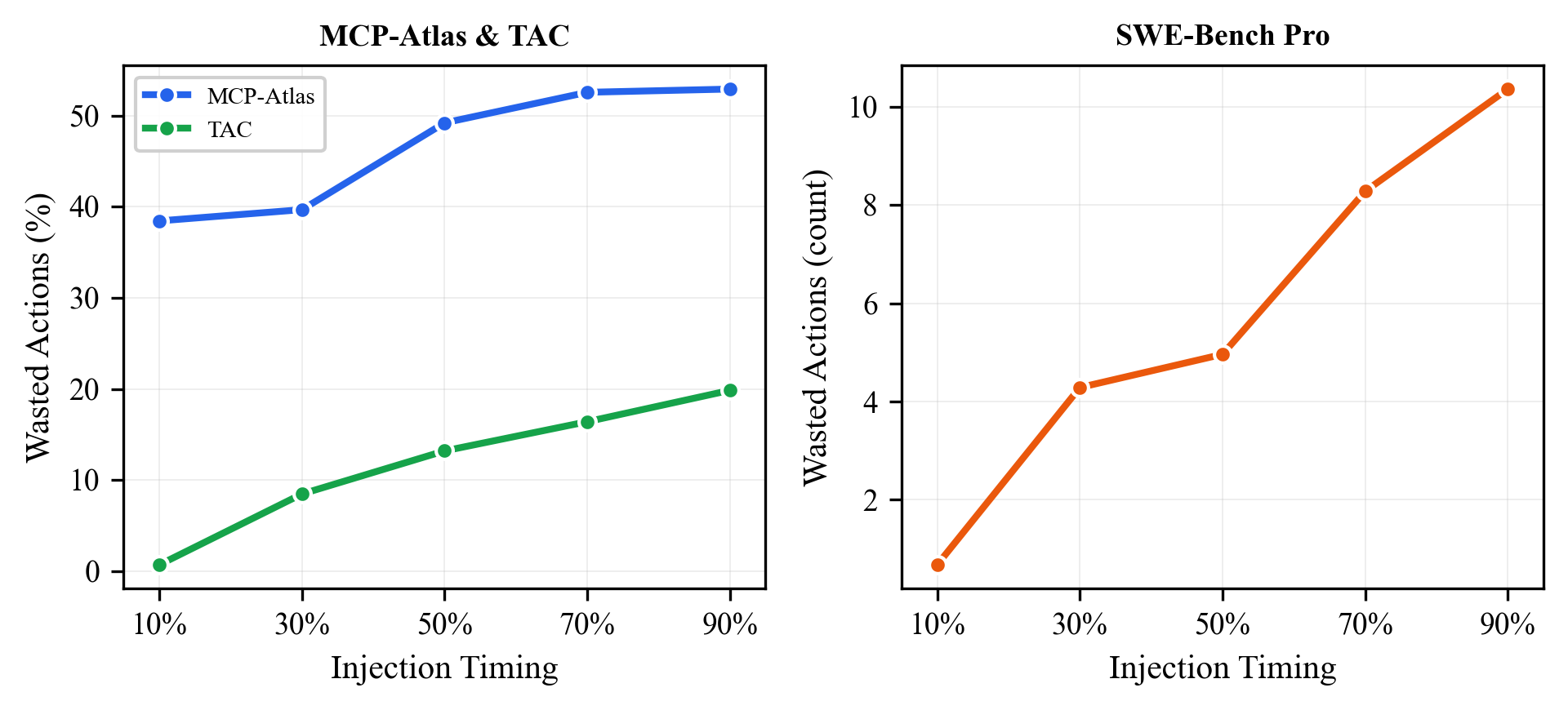}
  \caption{Wasted compute across all benchmarks. Left: MCP-Atlas and TheAgentCompany (fraction of pre-injection actions absent from the oracle trace). Right: SWE-Bench Pro (absolute action count; different scale due to longer trajectories). Later injection consistently produces more wasted work.}
  \label{fig:wasted_compute}
\end{figure}

\begin{table}[t]
  \caption{Cross-model Kendall's $\tau$ rank correlation. Left: TheAgentCompany only (3 models, filtered to complete-data pairs). Right: all benchmarks combined (4 models).}
  \label{tab:kendall}
  \centering
  \begin{subtable}[t]{0.45\textwidth}
    \centering
    \begin{tabular}{lccc}
      \toprule
      & Claude & Gemini & GPT \\
      \midrule
      Claude & 1.00 & 0.87 & 0.82 \\
      Gemini & 0.87 & 1.00 & 0.78 \\
      GPT & 0.82 & 0.78 & 1.00 \\
      \bottomrule
    \end{tabular}
    \caption{TheAgentCompany}
  \end{subtable}
  \hfill
  \begin{subtable}[t]{0.50\textwidth}
    \centering
    \begin{tabular}{lcccc}
      \toprule
      & Claude & DeepSeek & Gemini & GPT \\
      \midrule
      Claude & 1.00 & 0.45 & 0.67 & 0.43 \\
      DeepSeek & 0.45 & 1.00 & 0.47 & 0.34 \\
      Gemini & 0.67 & 0.47 & 1.00 & 0.39 \\
      GPT & 0.43 & 0.34 & 0.39 & 1.00 \\
      \bottomrule
    \end{tabular}
    \caption{All benchmarks combined}
  \end{subtable}
\end{table}

\section{Results}
\label{sec:results}

\subsection{VOI Curves Are Dimension-Dependent (H1)}

Figure~\ref{fig:voi_mcp} presents VOI curves on MCP-Atlas, the benchmark with the strongest timing signal ($n{=}5$--28 units per dimension; complete per-benchmark values in Appendix Table~\ref{tab:voi_full}). Clarification timing interacts strongly with information dimension, partially confirming H1 (the goal-front-loading prediction is strongly supported on MCP-Atlas; the context prediction is directionally consistent but underpowered at $n{=}17$ across two benchmarks):

\textbf{Goal.} Exhibits dramatic front-loading. Injection at 10\% of the trajectory recovers near-oracle performance (pass@3 of 0.78 vs.\ oracle 0.80, NC 0.40). The benefit decays steeply: Inj-30 = 0.50 (above NC), Inj-50 = 0.44 (above NC), Inj-70 = 0.39 ($\approx$ NC), Inj-90 = 0.39. By 70\% of the trajectory, goal injection no longer improves over no-clarification. This confirms that goal information has a narrow early window of high value.

\textbf{Input.} Shows gradual decline (0.46 at Inj-10, 0.36 at Inj-30 and Inj-50, 0.32 at Inj-70, 0.25 at Inj-90; NC = 0.33, oracle = 0.57). The early window (Inj-10 through Inj-50) recovers most of the oracle gap, but by Inj-70 input clarification falls to the NC baseline and by Inj-90 sits well below it. The flatter slope relative to goal confirms that input information is recoverable for longer than goal information, consistent with agents partially compensating through environment exploration, but this recovery window extends only through approximately 50\% of the trajectory.

\begin{figure}[t]
  \centering
  \includegraphics[width=0.80\linewidth]{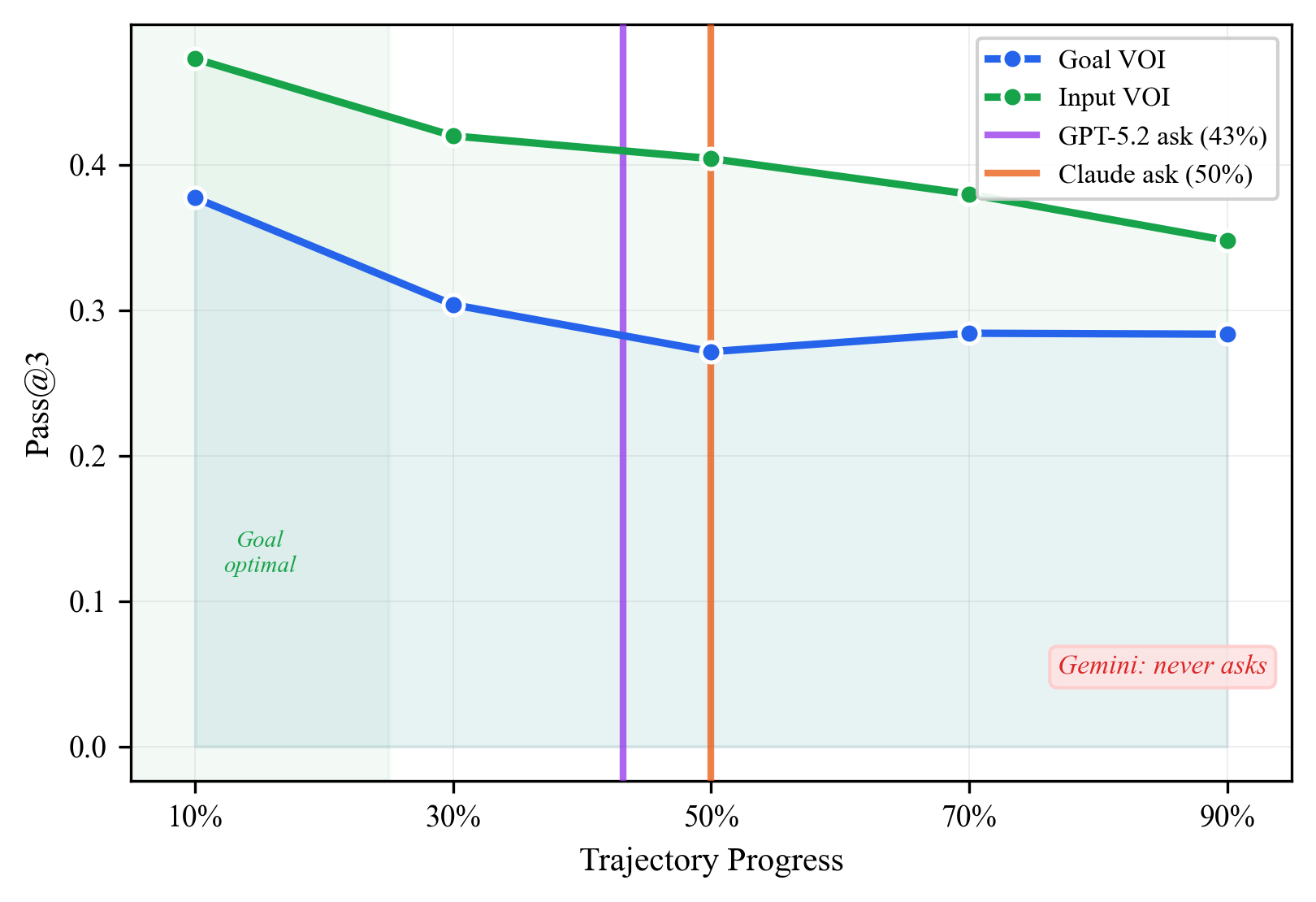}
  \caption{Natural ask timing overlaid on aggregate VOI curves for goal and input dimensions. Vertical lines indicate mean first-ask timing: GPT-5.2 (purple, 43\%) asks within the early-injection window for input but late for goal; Claude (orange, 50\%) asks at midpoint. Gemini never asks. The shaded region marks the optimal window for goal clarification (defined as the range of injection timings whose pass@3 is within 5 percentage points of the Inj-10 maximum).}
  \label{fig:natural_ask}
\end{figure}

\textbf{Constraint.} Constraint behavior depends on whether an oracle gap exists. On MCP-Atlas, Oracle and NC are identical (both 0.12), so the benchmark is uninformative for H2 (there is no information value to attenuate), and a flat injection curve (0.13--0.17 across timings) is consistent with both H2 and the null hypothesis. On SWE-Bench Pro, where the oracle gap is substantial (0.81 vs.\ 0.56 NC), constraint injection shows genuine declining benefit (0.81 at Inj-10, declining to 0.68 at Inj-90; Appendix Table~\ref{tab:voi_full}), supporting a weaker form of H2: reconciliation costs reduce constraint VOI relative to the oracle, but injected constraint information remains above the no-clarification baseline at all tested timings (0.68~$>$~0.56). The strong form of H2 (below-baseline disruption) is not supported by either benchmark. Constraint-type clarification should ideally be provided before execution begins rather than mid-trajectory.

\textbf{Context.} Evidence comes from two benchmarks with limited variant diversity: MCP-Atlas ($n{=}5$) shows strong early-injection benefit (0.80 at 10\% vs.\ 0.60 NC), and SWE-Bench Pro ($n{=}12$) is essentially flat across timings (0.92 at most injection points, with Oracle = 0.67 and NC = 0.75; the NC~$>$~Oracle inversion reflects sampling variability; see Appendix~\ref{sec:swe_limitations}). The combined sample ($n{=}17$) is too small for reliable conclusions about timing structure for context, though the MCP-Atlas signal is consistent with H1's front-loading prediction.

\textbf{Cross-benchmark patterns.} On TheAgentCompany, floor effects (oracle $\leq$29\%) attenuate timing signals. Goal injection conditions all sit within 4~pp of NC (0.12), consistent with sampling variability at $n{=}65$ rather than injection-induced disruption. Input shows mild benefit at intermediate timings (0.21--0.25 vs.\ 0.19 NC). Constraint is at floor (0\% across all conditions). SWE-Bench Pro results (Appendix~\ref{sec:per_benchmark_voi}; caveats in~\ref{sec:swe_limitations}) are directionally consistent. Aggregate curves appear in Appendix~\ref{sec:aggregate_interpret}.

\textbf{Point of no return.} We find that clarification benefit is continuous rather than threshold-based: only outcome-critical variants show a statistically significant recovery point, at 30\% of the trajectory ($p < 0.05$; Appendix~\ref{sec:ponr_detail}). Earlier is always better, but there is no sharp cutoff after which clarification becomes useless.

\subsection{Wasted Compute}

Wasted compute (fraction of pre-injection actions absent from the oracle trace, an upper bound on discarded work) increases steadily with injection delay across all three benchmarks (Figure~\ref{fig:wasted_compute}; Appendix~\ref{sec:wasted_compute_detail}). On TheAgentCompany, waste rises from 0.0\% at Inj-10 to 21.7\% at Inj-90. On MCP-Atlas, it ranges from 38.4\% to 52.9\% (higher baseline reflecting shorter trajectories). SWE-Bench Pro confirms the pattern: 0.7 to 10.4 wasted actions.

\subsection{Cross-Model Consistency}

Table~\ref{tab:kendall} reports Kendall's $\tau$ rank correlations between model pairs. On TheAgentCompany (3 models, balanced variant set), correlations range from 0.78 to 0.87, indicating strong agreement on which variants benefit from early clarification. On the combined dataset across all benchmarks (4 frontier models), Claude-Gemini achieves $\tau = 0.67$ and all model pairs exceed 0.34 ($p < 0.01$ for all pairs; see Appendix~\ref{sec:kendall_detail}). These results support the interpretation that timing effects are predominantly task-intrinsic: the same variants benefit from early injection regardless of which model executes them.

\subsection{Natural Ask Overlay}

Our natural-ask analysis of 300 TheAgentCompany sessions reveals stark differences in model clarification behavior (Appendix~\ref{sec:natural_ask_detail} provides per-model statistics and timing distributions). GPT-5.2 asks in 52\% of sessions with a mean first-ask timing of 43\% through the trajectory. Claude Sonnet 4.5 asks in 23\% of sessions at a mean timing of 50\%. Gemini 3 Flash never asks (0\% ask rate across all 100 sessions). These archetypes are consistent with LHAW's independent finding, under a different protocol, that GPT-5.2 over-clarifies with the lowest per-question efficiency while Gemini models under-clarify~\citep{pu2026lhaw}.

Overlaying these natural timings on the forced-injection VOI curves reveals timing alignment gaps (Figure~\ref{fig:natural_ask}). GPT-5.2's asking at 43\% is past the goal optimum (10\%) but within the input window; Claude's 50\% is suboptimal for goal but reasonable for input; Gemini forgoes all benefit. Notably, timing alignment alone does not predict success: Claude achieves higher per-session success (11\%) than GPT-5.2 (3\%) despite asking later and less often (Appendix~\ref{sec:natural_ask_detail}), suggesting that question quality may matter more than frequency, though this between-model comparison cannot isolate question quality from other model-level differences, and we treat it as a hypothesis rather than a controlled finding.

\section{Limitations}
\label{sec:limitations}

This study establishes the \emph{demand side} of clarification timing (how much does task performance benefit from receiving information at each trajectory point?) but does not address the \emph{supply side}: making agents recognize ambiguity and ask at the right moment. Building supply-side mechanisms on top of these VOI curves is an important next step. The strongest timing signals come from MCP-Atlas ($n{=}5$--28 per dimension); TheAgentCompany provides weaker signal due to floor effects, and SWE-Bench Pro has moderate per-cell sizes ($n{=}12$--31). The natural-ask protocol covers only TheAgentCompany (300 sessions). A behavioral confound exists between protocols: forced injection disables \texttt{ask\_user} while natural-ask enables it, so agents may plan differently when they know they cannot ask; the forced-injection VOI curves should be interpreted as upper bounds on information value.

\section{Conclusion}

We present the first empirical VOI curves for clarification timing in long-horizon agent workflows. Timing effects are strongly dimension-dependent: goal clarification is front-loaded (${\leq}$10\% window), input degrades gradually (recoverable through ${\sim}$50\%), and constraint benefit depends on the oracle gap, with cross-model consistency confirming these profiles are task-intrinsic. No current frontier model asks within the optimal window. These findings yield two actionable guidelines: (1)~agents should validate goals within the first few actions; (2)~input queries can be deferred slightly longer but should be raised within the first half of the trajectory. More broadly, the forced-injection methodology generalizes to any agent benchmark with underspecified tasks and provides the empirical demand curves that existing theoretical frameworks~\citep{howard1966information, dong2026voi, kobalczyk2025active} require but have lacked. Building timing-aware clarification policies~\citep{sen2026chronos} on top of these curves is a natural next step.

\begin{ack}
We thank the LHAW, MCP-Atlas, TheAgentCompany, and SWE-Bench Pro teams for their open-source benchmarks. We also thank our colleagues at PricewaterhouseCoopers for valuable feedback on earlier drafts.
\end{ack}

\bibliographystyle{plainnat}
\bibliography{references}

\appendix

\section{Extended Results}
\label{sec:extended_results}

\subsection{Per-Benchmark VOI Curves}
\label{sec:per_benchmark_voi}

\begin{figure}[h]
  \centering
  \includegraphics[width=\linewidth]{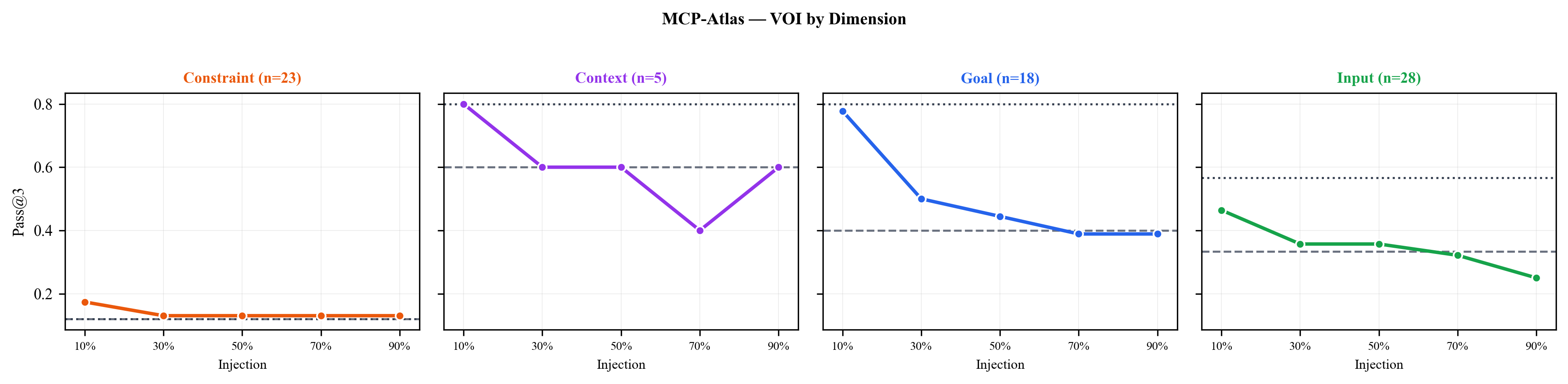}
  \caption{MCP-Atlas VOI curves by dimension (expanded view with per-model error indicators; cf.\ the pooled summary in main text Figure~\ref{fig:voi_mcp}). Goal shows dramatic front-loading (0.78 at 10\% vs.\ 0.40 NC), input shows gradual decline, constraint is flat near zero, and context (limited $n{=}5$) shows strong early benefit. Dashed: no-clarification baseline; dotted: oracle.}
  \label{fig:appendix_voi_mcp}
\end{figure}

\begin{figure}[h]
  \centering
  \includegraphics[width=\linewidth]{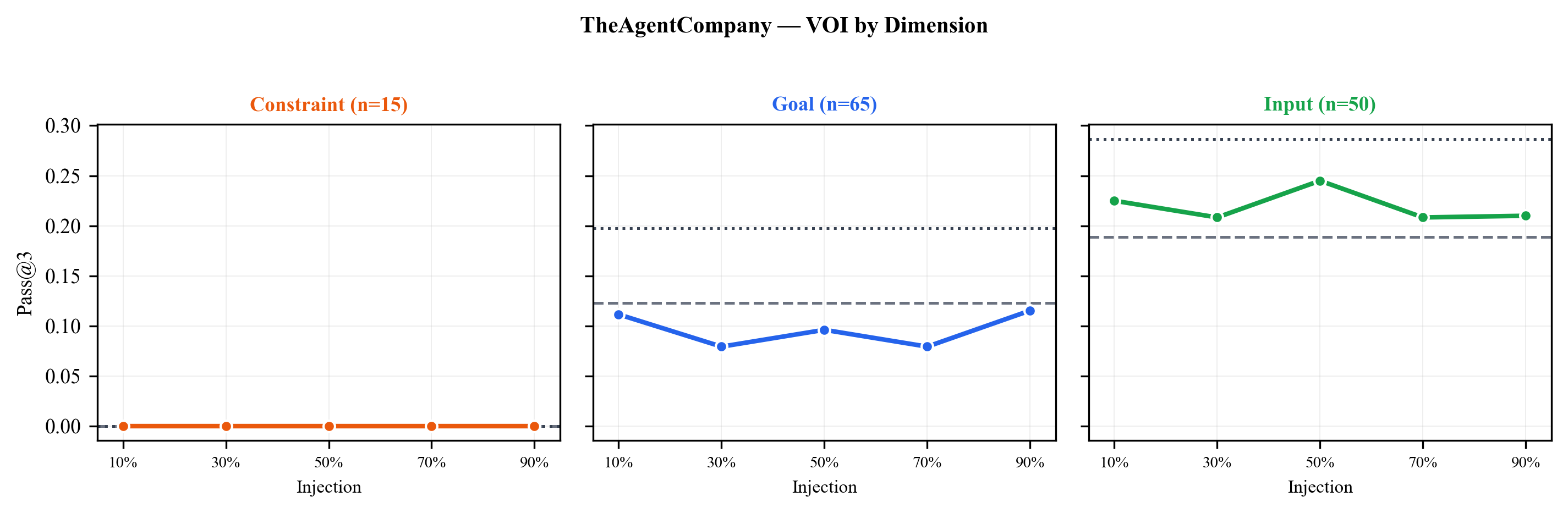}
  \caption{TheAgentCompany VOI curves by dimension. Success rates are substantially lower than MCP-Atlas (oracle $\leq$29\% for non-constraint dimensions). Goal shows no reliable timing effect (all conditions within 4 percentage points of baseline). Input shows a mild benefit at intermediate timings (21--25\% at Inj-30/50/70 vs.\ 19\% NC). Constraint is at floor (0\% across all conditions).}
  \label{fig:appendix_voi_tac}
\end{figure}

\begin{figure}[h]
  \centering
  \includegraphics[width=\linewidth]{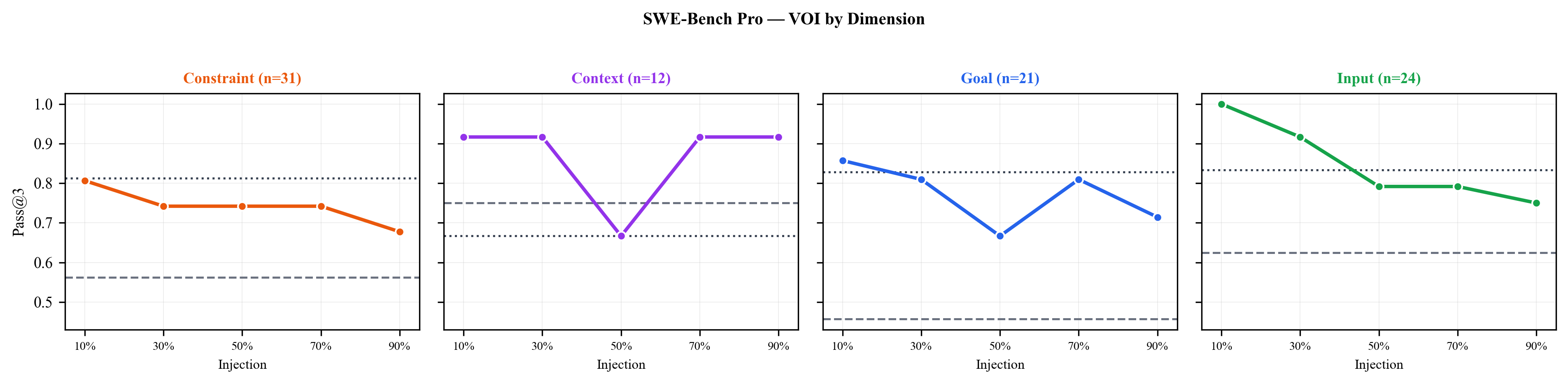}
  \caption{SWE-Bench Pro VOI curves by dimension, pooled across three models ($n{=}12$--31 per injection cell). All four dimensions show injection benefit over no-clarification, with input and goal exhibiting the clearest declining VOI pattern. Context has limited variant diversity ($n{=}12$).}
  \label{fig:appendix_voi_swe}
\end{figure}

Table~\ref{tab:voi_full} presents the complete pass@3 values for all (benchmark, dimension, condition) cells. Figures~\ref{fig:appendix_voi_mcp}--\ref{fig:appendix_voi_swe} visualize these as per-benchmark VOI curves.

\begin{table}[h]
  \caption{Complete pass@3 results by benchmark, dimension, and condition. $n$ denotes injection-condition (variant, model) units per dimension; only pairs for which all five injection conditions and the oracle baseline completed successfully are included (pairs with failures in any condition are excluded). For multi-dimension variants (Table~\ref{tab:variant_breakdown}), each pair is attributed to its primary dimension only. Oracle and NC cells may include up to 4 additional baseline-only units (per-cell $n$ for Oracle/NC is therefore $n + k$ where $k \in [0, 4]$; exact counts are available in the released code). TheAgentCompany uses the 3-model subset (see Section~\ref{sec:benchmarks}).}
  \label{tab:voi_full}
  \centering
  \small
  \begin{tabular}{llccccccc}
    \toprule
    Benchmark & Dimension & Oracle & Inj-10 & Inj-30 & Inj-50 & Inj-70 & Inj-90 & NC \\
    \midrule
    \multirow{4}{*}{MCP-Atlas} 
    & Goal ($n{=}18$) & 0.80 & 0.78 & 0.50 & 0.44 & 0.39 & 0.39 & 0.40 \\
    & Constraint ($n{=}23$) & 0.12 & 0.17 & 0.13 & 0.13 & 0.13 & 0.13 & 0.12 \\
    & Input ($n{=}28$) & 0.57 & 0.46 & 0.36 & 0.36 & 0.32 & 0.25 & 0.33 \\
    & Context ($n{=}5$) & 0.80 & 0.80 & 0.60 & 0.60 & 0.40 & 0.60 & 0.60 \\
    \midrule
    \multirow{3}{*}{TAC}
    & Goal ($n{=}65$) & 0.20 & 0.11 & 0.08 & 0.10 & 0.08 & 0.12 & 0.12 \\
    & Constraint ($n{=}15$) & 0.00 & 0.00 & 0.00 & 0.00 & 0.00 & 0.00 & 0.00 \\
    & Input ($n{=}50$) & 0.29 & 0.23 & 0.21 & 0.25 & 0.21 & 0.21 & 0.19 \\
    \midrule
    \multirow{4}{*}{SWE-Bench Pro}
    & Goal ($n{=}21$) & 0.83 & 0.86 & 0.81 & 0.67 & 0.81 & 0.71 & 0.46 \\
    & Constraint ($n{=}31$) & 0.81 & 0.81 & 0.74 & 0.74 & 0.74 & 0.68 & 0.56 \\
    & Input ($n{=}24$) & 0.83 & 1.00 & 0.92 & 0.79 & 0.79 & 0.75 & 0.63 \\
    & Context ($n{=}12$) & 0.67 & 0.92 & 0.92 & 0.67 & 0.92 & 0.92 & 0.75 \\
    \bottomrule
  \end{tabular}
\end{table}

\textbf{Benchmark-specific interpretation.} The three benchmarks represent a spectrum of task characteristics that explain their distinct VOI patterns:

\begin{itemize}
  \item \textbf{MCP-Atlas} tasks are short (6--20 actions), tool-focused, and typically have a single correct approach. This creates high commitment rates: choosing the wrong tool chain in the first few actions wastes most of the trajectory. Goal clarification is therefore extremely front-loaded.
  
  \item \textbf{TheAgentCompany} tasks are longer (6--49 actions) and involve multiple interacting systems (web, database, file system). The lower baselines reflect genuine task difficulty; even oracle-informed agents achieve only 20--29\% pass@3 on non-constraint dimensions. The weak timing signal for goal suggests that these tasks have inherently high stochasticity: success depends on factors beyond information completeness (e.g., correct navigation of web interfaces).
  
  \item \textbf{SWE-Bench Pro} tasks vary enormously in difficulty (1--121 actions). Input shows a gradual decline from 1.00 (Inj-10) to 0.75 (Inj-90), consistent with code-generation patterns: once an agent commits to an implementation approach (choosing file locations, function signatures, test strategies), switching costs increase steadily.
\end{itemize}

\subsection{Natural Ask Timing Details}
\label{sec:natural_ask_detail}

The 300 natural-ask sessions span seven TheAgentCompany base tasks. Table~\ref{tab:natural_ask_stats} summarizes per-model ask behavior.

\begin{table}[h]
  \caption{Natural ask behavior by model (100 sessions each). Ask rate is the fraction of sessions with $\geq$1 ask call. Mean timing is the trajectory-relative position of the first ask.}
  \label{tab:natural_ask_stats}
  \centering
  \begin{tabular}{lcccc}
    \toprule
    Model & Ask Rate & Total Calls & Mean Timing & Median Timing \\
    \midrule
    GPT-5.2 & 52\% & 89 & 43\% & 50\% \\
    Claude Sonnet 4.5 & 23\% & 23 & 50\% & 50\% \\
    Gemini 3 Flash & 0\% & 0 & -- & -- \\
    \bottomrule
  \end{tabular}
\end{table}

GPT-5.2 averages 1.71 ask calls per asking session (max 5), indicating a pattern of iterative clarification. Claude always asks exactly once when it asks at all, suggesting a more deliberate single-query strategy. The distribution of first-ask timings is concentrated: 72\% of all first-ask events occur between 25\% and 50\% of the trajectory.

\begin{figure}[h]
  \centering
  \includegraphics[width=\linewidth]{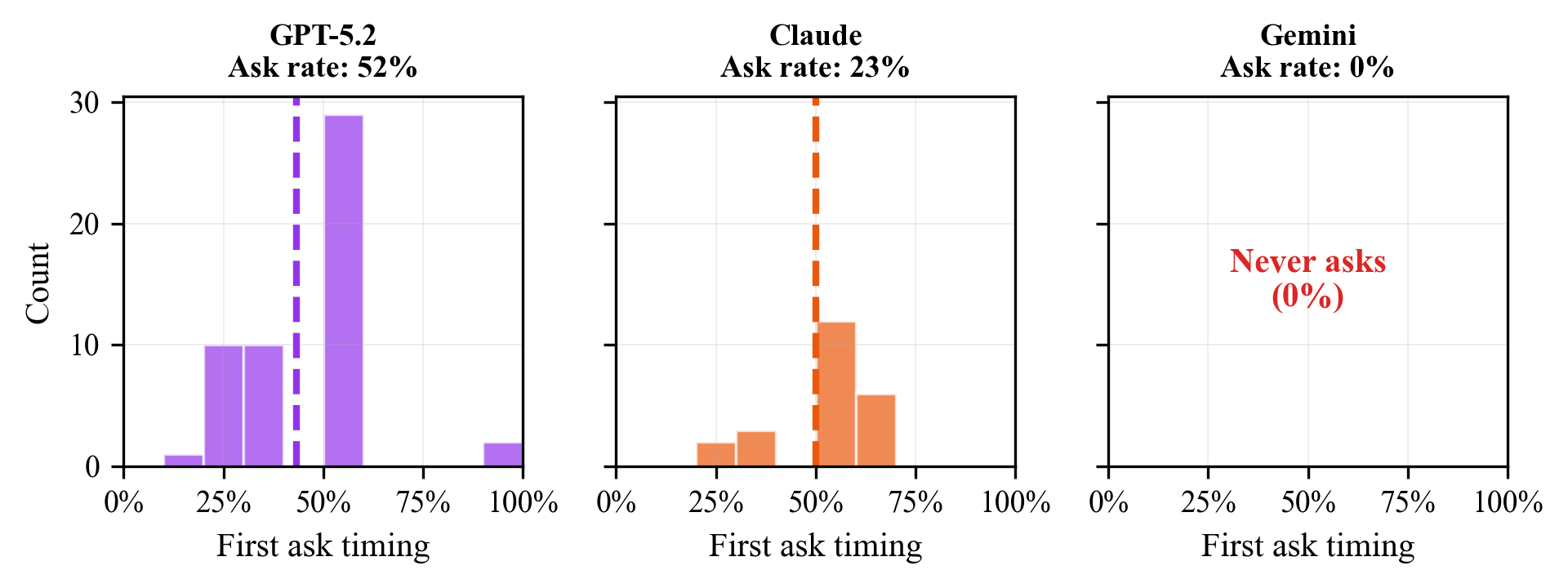}
  \caption{Distribution of first-ask timing by model. GPT-5.2 asks frequently (52\% of sessions) with timing concentrated between 25--50\% of the trajectory. Claude asks selectively (23\%) with a tight cluster around 50\%. Gemini never asks. Dashed lines indicate mean first-ask timing.}
  \label{fig:appendix_ask_timing_dist}
\end{figure}

\begin{figure}[h]
  \centering
  \includegraphics[width=\linewidth]{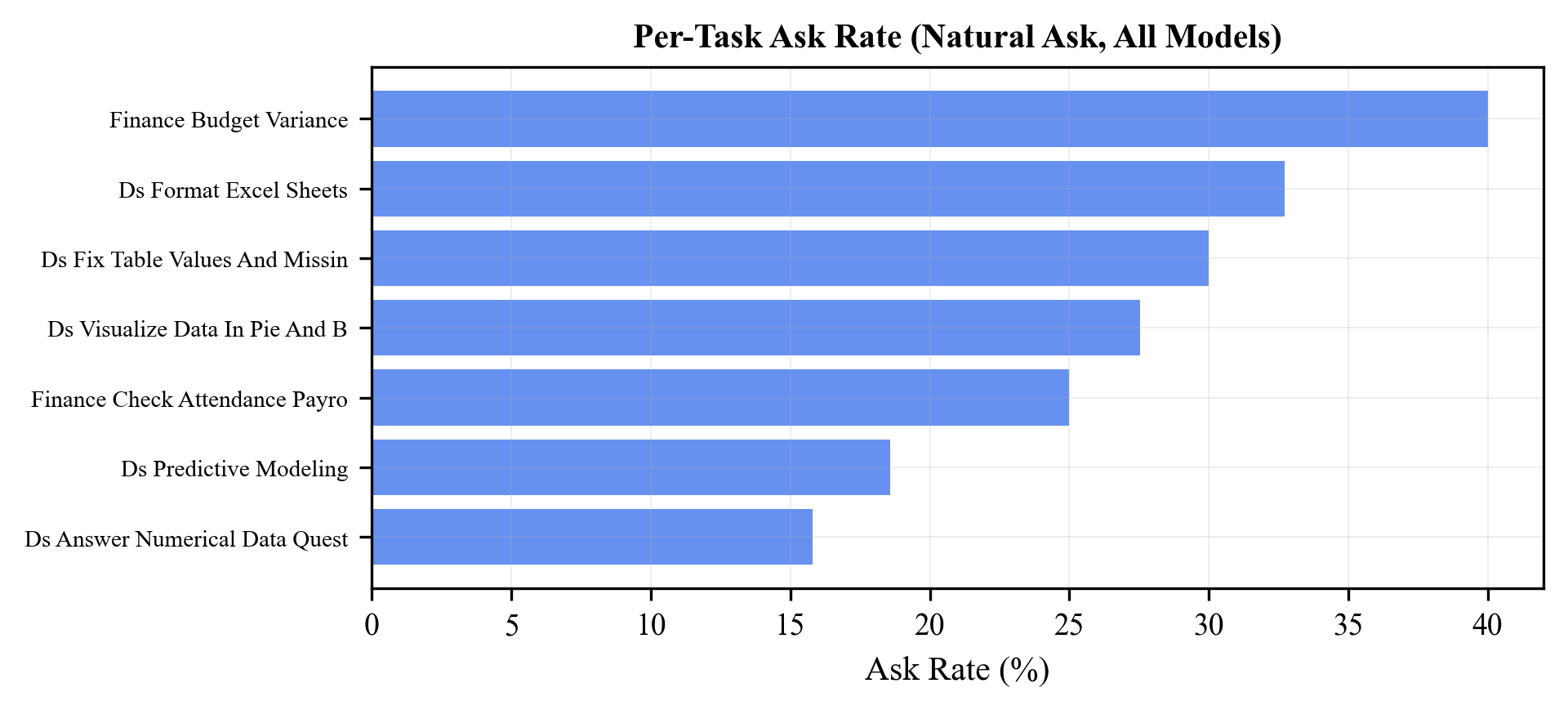}
  \caption{Per-task ask rate across all models. Task complexity appears to influence asking propensity independently of model architecture, with data-analysis tasks eliciting higher ask rates than simpler administrative tasks.}
  \label{fig:appendix_ask_per_task}
\end{figure}

\textbf{Qualitative analysis of ask content.} GPT-5.2's questions tend to be specific and actionable (e.g., ``Should I use the FY25 Q1 data or the calendar year Q1?''), while Claude's questions are broader (e.g., ``Could you clarify what format you'd like the output in?''). Despite asking far more often, GPT-5.2 achieves substantially lower per-session success (3\%) than Claude (11\%), indicating that asking frequency does not predict task success. Claude's selective single-question strategy outperforms GPT-5.2's iterative approach, suggesting that question quality and selectivity may matter more than volume, though we cannot isolate this from other model-level differences. This is consistent with the VOI framework: later and repeated questions yield diminishing returns, and poorly targeted questions may provide no actionable benefit even when asked at an appropriate trajectory point.

\subsection{Wasted Compute by Benchmark}
\label{sec:wasted_compute_detail}

Table~\ref{tab:wasted_detail} reports mean wasted compute fractions by benchmark and injection timing. On TheAgentCompany, wasted compute follows a clean linear pattern reflecting the longer trajectories where waste accumulates gradually. On MCP-Atlas, the higher baseline (38\% even at injection-10) reflects shorter trajectories where even a single early action pursuing the wrong path counts as waste relative to the oracle trace.

\begin{table}[h]
  \caption{Wasted compute by benchmark and injection timing. MCP-Atlas and TheAgentCompany report the fraction of pre-injection actions absent from the oracle trace; SWE-Bench Pro reports absolute action counts due to longer, variable-length trajectories.}
  \label{tab:wasted_detail}
  \centering
  \begin{tabular}{lccccc}
    \toprule
    Benchmark & Inj-10 & Inj-30 & Inj-50 & Inj-70 & Inj-90 \\
    \midrule
    TheAgentCompany (\%) & 0.0 & 9.6 & 13.7 & 17.6 & 21.7 \\
    MCP-Atlas (\%) & 38.4 & 39.7 & 49.2 & 52.6 & 52.9 \\
    SWE-Bench Pro (actions) & 0.7 & 4.3 & 5.0 & 8.3 & 10.4 \\
    \bottomrule
  \end{tabular}
\end{table}

The MCP-Atlas pattern of high initial waste (38\% at injection-10) that plateaus at 53\% reflects a structural property of short tool-calling trajectories: even early injection requires the agent to restart its approach, and the first 1--2 actions are often already committed. In contrast, TheAgentCompany's linear growth from 0\% reflects that the first few actions in enterprise workflows (opening a browser, navigating to a page) are typically approach-independent and not wasted regardless of the eventual clarification.

\subsection{Cross-Model Correlation Details}
\label{sec:kendall_detail}

\begin{figure}[h]
  \centering
  \begin{minipage}{0.48\textwidth}
    \centering
    \includegraphics[width=\linewidth]{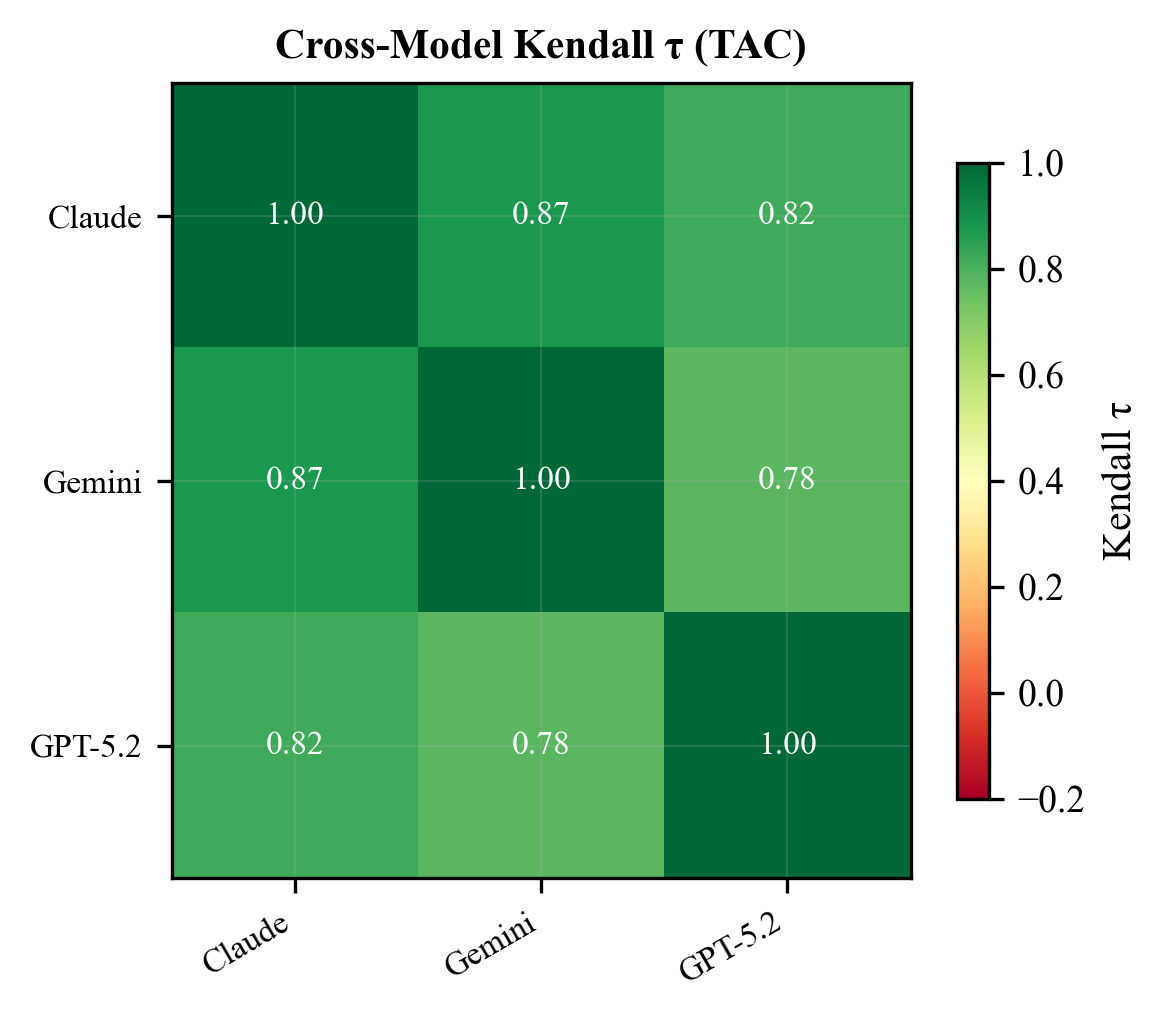}
  \end{minipage}
  \hfill
  \begin{minipage}{0.48\textwidth}
    \centering
    \includegraphics[width=\linewidth]{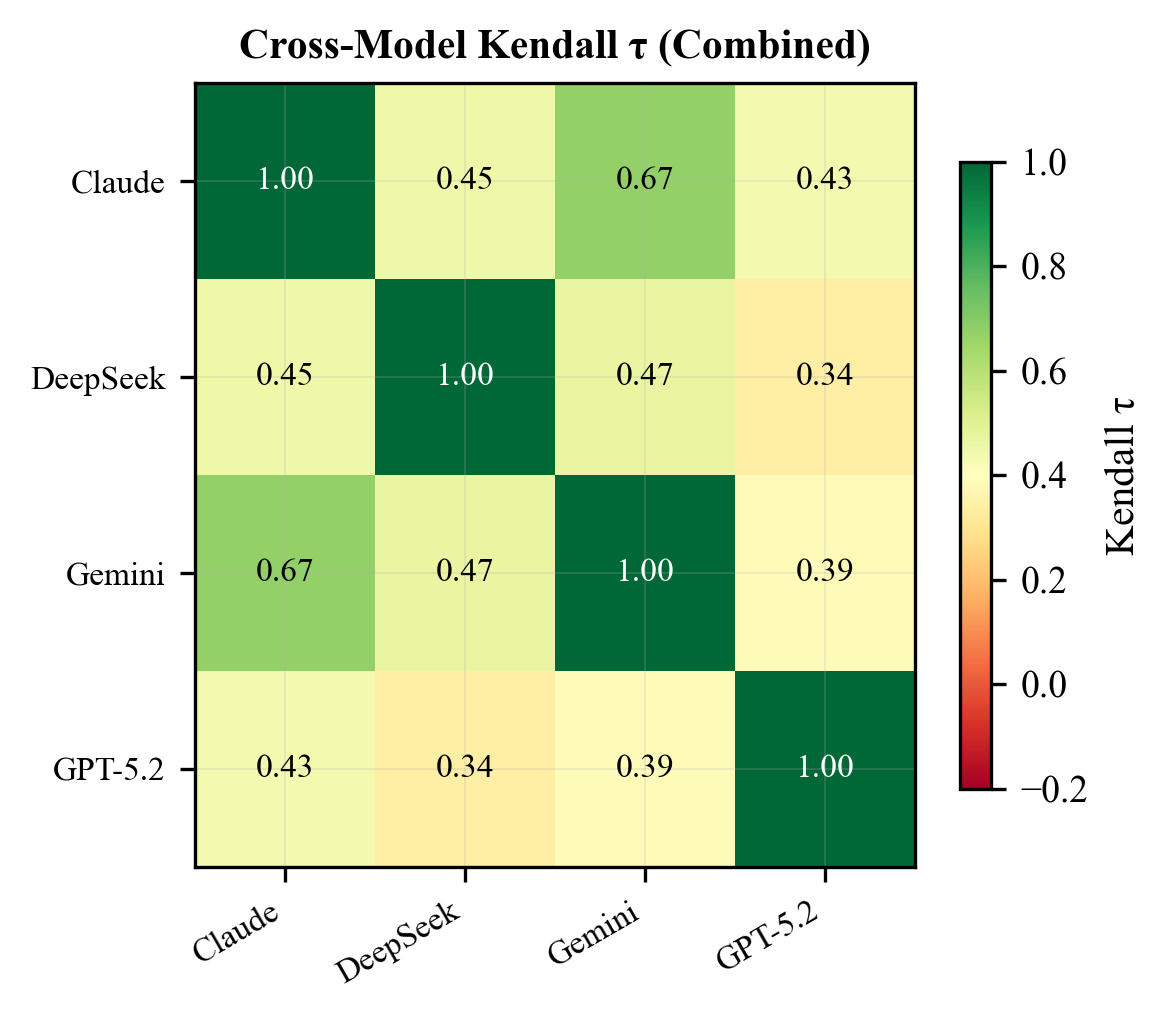}
  \end{minipage}
  \caption{Kendall $\tau$ correlation heatmaps. Left: TheAgentCompany only (3 models, balanced variant set). Right: all benchmarks combined (4 frontier models). All $p$-values $< 0.01$ except where noted.}
  \label{fig:appendix_kendall}
\end{figure}

Table~\ref{tab:kendall_pvalues} reports the $p$-values for all model-pair correlations. For the combined dataset, all 4-model pairs achieve significance at $p < 0.01$. For the TAC-only analysis, all pairs are significant at $p < 10^{-5}$, reflecting the larger number of comparable variant-timing combinations (each model runs all 30 TAC variants across 5 injection conditions).

\begin{table}[h]
  \caption{Kendall $\tau$ $p$-values for all model pairs. Combined: all benchmarks (4 models). TAC: TheAgentCompany only (3 models, filtered to complete-data pairs).}
  \label{tab:kendall_pvalues}
  \centering
  \small
  \begin{subtable}[t]{0.47\textwidth}
    \centering
    \begin{tabular}{lccc}
      \toprule
      & Claude & Gemini & GPT \\
      \midrule
      Claude & -- & $< 10^{-9}$ & $< 10^{-4}$ \\
      Gemini & & -- & $< 10^{-3}$ \\
      GPT & & & -- \\
      \bottomrule
    \end{tabular}
    \caption{TAC $p$-values}
  \end{subtable}
  \hfill
  \begin{subtable}[t]{0.50\textwidth}
    \centering
    \begin{tabular}{lcccc}
      \toprule
      & Claude & DeepSeek & Gemini & GPT \\
      \midrule
      Claude & -- & $< 10^{-3}$ & $< 10^{-9}$ & $< 10^{-4}$ \\
      DeepSeek & & -- & $< 10^{-3}$ & $< 0.01$ \\
      Gemini & & & -- & $< 10^{-3}$ \\
      GPT & & & & -- \\
      \bottomrule
    \end{tabular}
    \caption{Combined $p$-values}
  \end{subtable}
\end{table}

The higher TAC correlations (0.78--0.87) compared to the combined dataset (0.34--0.67) likely reflect two factors: (1) the TAC subset uses exactly the same variant set and injection protocol across all three models (ensuring apple-to-apple comparison), and (2) the combined dataset includes MCP-Atlas with 4 models (introducing DeepSeek which is only available on that benchmark) and SWE-Bench Pro where high overall pass rates compress the VOI range, attenuating rank correlations.

\subsection{Point of No Return Analysis}
\label{sec:ponr_detail}

\begin{figure}[h]
  \centering
  \includegraphics[width=0.7\linewidth]{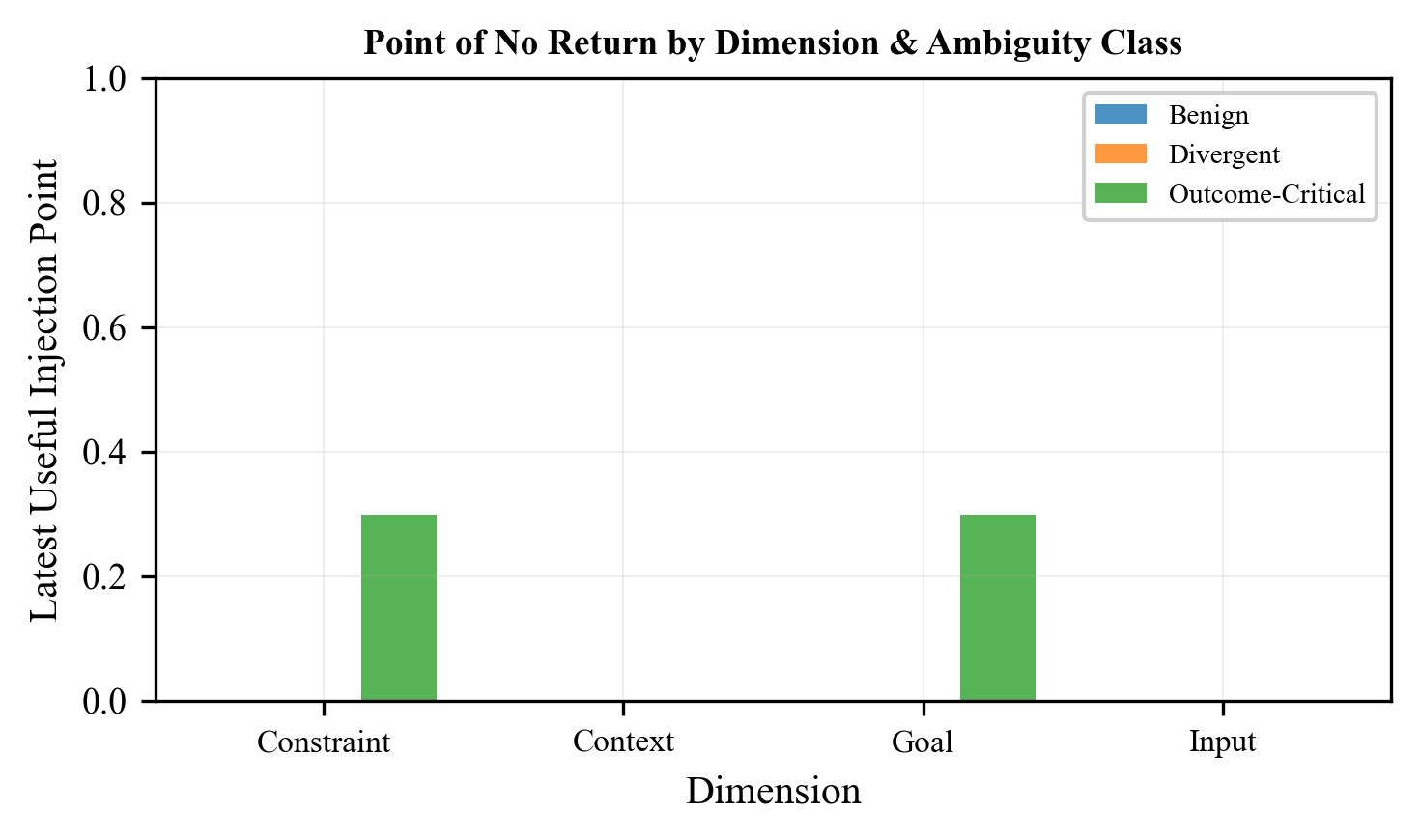}
  \caption{Latest useful injection point by dimension and ambiguity class. Only outcome-critical variants for goal and constraint dimensions show a statistically significant recovery point (at 30\% of trajectory). Blank bars indicate no significant point of no return was detected.}
  \label{fig:appendix_ponr}
\end{figure}

Table~\ref{tab:ponr} reports the point-of-no-return analysis. We define the ``latest useful injection point'' as the latest timing at which pass@3 significantly exceeds the no-clarification baseline (one-sided permutation test with 10,000 permutations, $p < 0.05$, Bonferroni-corrected for 5 timing comparisons per cell).

\begin{table}[h]
  \caption{Point of no return: latest injection timing with significant improvement over no-clarification ($p < 0.05$, Bonferroni-corrected). ``--'' indicates no timing achieves significance.}
  \label{tab:ponr}
  \centering
  \begin{tabular}{lccc}
    \toprule
    Dimension & Outcome-Critical & Divergent & Benign \\
    \midrule
    Goal & 30\% & -- & -- \\
    Constraint & 30\% & -- & -- \\
    Input & -- & -- & -- \\
    Context & -- & -- & -- \\
    \bottomrule
  \end{tabular}
\end{table}

The absence of significant points for input and context dimensions is consistent with the VOI curves: input shows gradual decline (never sharply crossing the baseline), and context has too few samples ($n{=}5$) for statistical power. For goal and constraint, the 30\% cutoff aligns with the commitment model's prediction that these dimensions have steep early commitment rates.

\subsection{Qualitative Trajectory Analysis}
\label{sec:qualitative}

To illustrate how timing interacts with agent behavior, we present representative trajectory patterns for each dimension:

\textbf{Goal (MCP-Atlas, early injection helps).} A task requires exporting data ``in a format suitable for spreadsheet analysis.'' Without clarification, all three models default to JSON output. With goal clarification at 10\% (``the user wants CSV format''), the agent correctly outputs CSV from the start. At 50\%, two of three models have already committed to JSON parsing pipelines and must discard 3--5 actions to restart. At 90\%, all models have completed the task in the wrong format.

\textbf{Input (MCP-Atlas, gradual decline).} A task requires querying ``the customer database'' but doesn't specify which table contains customer records. Without clarification, agents explore the schema (listing tables, examining columns), eventually finding the correct table through trial and error. This exploration is ``wasted'' relative to the oracle trace but is informationally productive, as it narrows the search space. Early injection eliminates this exploration; late injection still saves the agent from querying the wrong table at the final step.

\textbf{Constraint (TAC, disruptive injection).} A task asks for a budget report but omits that ``all figures must be in thousands.'' Without this constraint, agents produce correct reports in raw units. When the constraint is injected at 50\%, agents attempt to retroactively scale existing outputs, sometimes introducing rounding errors or formatting inconsistencies that reduce checkpoint scores below the unconstrained baseline.

\subsection{SWE-Bench Pro Limitations}
\label{sec:swe_limitations}

SWE-Bench Pro results warrant additional caveats beyond those noted in Section~\ref{sec:limitations}:

\begin{enumerate}
  \item \textbf{Moderate sample sizes.} Pooling across three models yields $n{=}12$--31 (variant, model) units per injection cell depending on dimension. Per-model estimates remain noisy, but the pooled values in Table~\ref{tab:voi_full} are more stable than single-model estimates.
  
  \item \textbf{Model coverage.} Three models (Gemini 3 Flash, GPT-5.2, Claude Sonnet 4.5) have both injection and baseline data, enabling within-model comparisons. DeepSeek V3.2 is excluded from SWE-Bench Pro.
  
  \item \textbf{Context dimension.} Only 4 context-type variants are available in LHAW's SWE-Bench Pro subset, yielding $n{=}12$ units when pooled across models. The resulting estimates should be interpreted with caution given the limited variant diversity.
\end{enumerate}

Additionally, multiple SWE-Bench Pro cells exhibit values above the Oracle baseline: Goal Inj-10 (0.86 vs.\ Oracle 0.83), Input Inj-10 (1.00 vs.\ 0.83), Input Inj-30 (0.92 vs.\ 0.83), Context Inj-10 (0.92 vs.\ 0.67), and Context NC (0.75 vs.\ 0.67). At sample sizes of $n{=}12$--31 per cell, these inversions are consistent with sampling variability. An additional contributing factor is that the injection messages provide explicit, structured guidance (e.g., ``the target format is CSV'') that may be more salient than the same information embedded in the original task prompt; this is a known artifact of forced-injection protocols. None of these differences is statistically significant. We retain Oracle as the conceptual upper bound while noting that small-sample noise can produce localized inversions.

Despite these limitations, SWE-Bench Pro contributes to the aggregate picture by confirming that code-generation tasks exhibit steep commitment rates (consistent with the goal-dimension pattern on MCP-Atlas) and that oracle information substantially improves performance (0.81--0.83 oracle vs.\ 0.46--0.63 no-clarification for goal, constraint, and input dimensions across three models; see Table~\ref{tab:voi_full}).

\subsection{Total Runs and Computational Cost}
\label{sec:compute_cost}

Table~\ref{tab:compute} summarizes the total experimental scope.

\begin{table}[h]
  \caption{Total experimental trials by benchmark and protocol. Each forced-injection trial corresponds to one (variant, model, condition, trial-index) execution. Natural-ask sessions are single-trial.}
  \label{tab:compute}
  \centering
  \begin{tabular}{lccc}
    \toprule
    Benchmark & Forced Injection Trials & Natural Ask Sessions & Total Trials \\
    \midrule
    MCP-Atlas & 3,024 & -- & 3,024 \\
    TheAgentCompany & 1,890 & 300 & 2,190 \\
    SWE-Bench Pro & 1,134 & -- & 1,134 \\
    \midrule
    \textbf{Total} & \textbf{6,048} & \textbf{300} & \textbf{6,348} \\
    \bottomrule
  \end{tabular}
\end{table}

\subsection{Aggregate VOI Interpretation}
\label{sec:aggregate_interpret}

\begin{figure}[h]
  \centering
  \includegraphics[width=\linewidth]{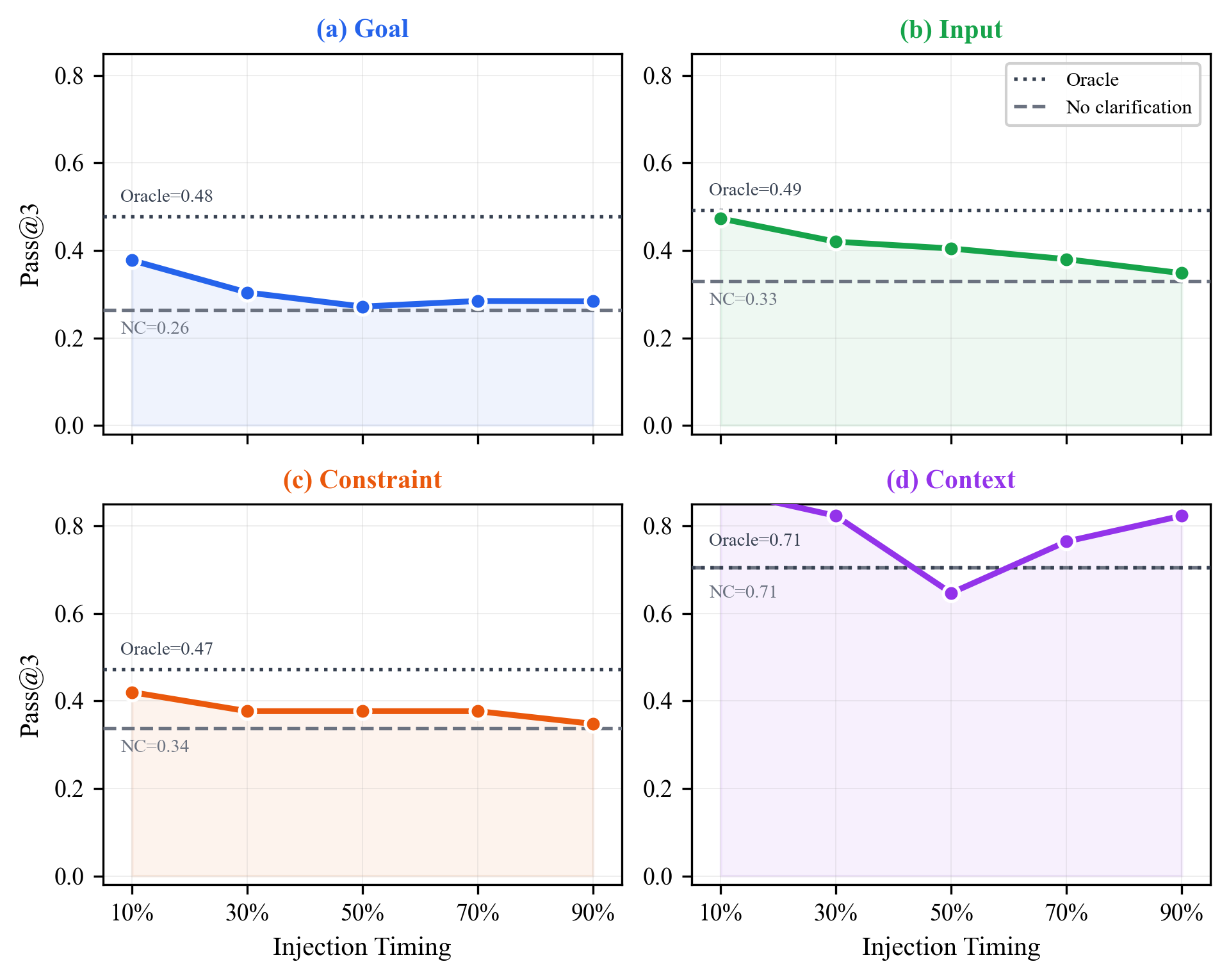}
  \caption{Aggregate VOI curves by information dimension (all benchmarks combined). Goal clarification is front-loaded, input shows gradual decline, and constraint injection provides only marginal benefit over no-clarification. Dashed lines indicate oracle (upper) and no-clarification (lower) baselines. See main text Figure~\ref{fig:voi_mcp} for the per-benchmark MCP-Atlas curves that drive the strongest signals.}
  \label{fig:voi_aggregate}
\end{figure}

The aggregate curves in Figure~\ref{fig:voi_aggregate} pool across benchmarks with substantially different baseline performance levels, weighting each (variant, model) unit equally. Because TAC contributes the majority of goal units (63\%) with near-floor baselines, the aggregate attenuates the stronger timing signals visible in the per-benchmark curves (Figures~\ref{fig:appendix_voi_mcp}--\ref{fig:appendix_voi_swe}). The per-benchmark breakdowns should be consulted alongside this summary view.

\subsection{Dataset Details}
\label{sec:dataset_details}

Table~\ref{tab:variant_breakdown} provides the full cross-tabulation of the 84 task variants used in this study, broken down by benchmark, information dimension(s), and ambiguity class. The subset is drawn from a larger pool of 101 variants in LHAW~\citep{pu2026lhaw}; 17 SWE-Bench Pro variants are excluded to manage computational costs. Selection was stratified to preserve dimension balance: excluded variants do not differ systematically from retained ones on dimension distribution or ambiguity class (see Section~\ref{sec:limitations}).

\begin{table}[h]
  \caption{Task variant breakdown by benchmark, dimension, and ambiguity class. Multi-dimension variants (e.g., ``goal+input'') have multiple segments removed simultaneously. Totals: 36 MCP-Atlas, 30 TheAgentCompany, 18 SWE-Bench Pro.}
  \label{tab:variant_breakdown}
  \centering
  \small
  \begin{tabular}{llcccc}
    \toprule
    Benchmark & Dimension(s) & Outcome-Crit. & Divergent & Benign & Total \\
    \midrule
    \multirow{5}{*}{MCP-Atlas}
    & goal & 6 & 1 & 0 & 7 \\
    & input & 5 & 4 & 0 & 9 \\
    & constraint & 5 & 4 & 2 & 11 \\
    & context / context+goal & 0 & 1 & 2 & 3 \\
    & multi (goal+input, constr.+input, constr.+goal) & 2 & 1 & 3 & 6 \\
    \midrule
    \multirow{4}{*}{TAC}
    & goal & 6 & 2 & 3 & 11 \\
    & input & 2 & 4 & 0 & 6 \\
    & constraint / constr.+goal & 2 & 0 & 1 & 3 \\
    & multi (goal+input, constr.+input) & 4 & 5 & 1 & 10 \\
    \midrule
    \multirow{4}{*}{SWE-Bench Pro}
    & goal+input / input & 1 & 2 & 2 & 5 \\
    & constraint / constr.+goal & 2 & 2 & 2 & 6 \\
    & context / context+goal & 0 & 1 & 2 & 3 \\
    & constr.+input / constr.+context & 2 & 1 & 1 & 4 \\
    \midrule
    \multicolumn{2}{l}{\textbf{Total}} & \textbf{37} & \textbf{28} & \textbf{19} & \textbf{84} \\
    \bottomrule
  \end{tabular}
\end{table}

\subsection{Injection Protocol Details}
\label{sec:injection_details}

\subsubsection{Injection Message Construction}

Injection messages are constructed from a template system that maps each (dimension, subdimension) pair to a natural-language message pattern. Table~\ref{tab:templates} shows representative templates. For variants with multiple removed segments, the first segment uses the full template and subsequent segments are appended with connective phrases (``And \{value\}.'' or ``Also, \{value\}.'').

\begin{table}[h]
  \caption{Representative injection message templates by dimension and subdimension. The \texttt{\{value\}} placeholder is filled with the ground-truth content from the LHAW variant specification.}
  \label{tab:templates}
  \centering
  \small
  \begin{tabular}{lll}
    \toprule
    Dimension & Subdimension & Template \\
    \midrule
    goal & target & ``By the way, I should clarify: I'm specifically looking for \{value\}.'' \\
    goal & format & ``By the way, please give me the result in \{value\}.'' \\
    input & source & ``By the way, you can find the data in \{value\}.'' \\
    input & location & ``By the way, it's located at \{value\}.'' \\
    constraint & temporal & ``By the way, I should have mentioned: I'm looking at \{value\}.'' \\
    constraint & selection & ``I should mention, only include those that are \{value\}.'' \\
    context & background & ``By the way, for context: \{value\}.'' \\
    context & domain knowledge & ``For context, \{value\}.'' \\
    \midrule
    \multicolumn{2}{l}{(default fallback)} & ``By the way, I should have mentioned: \{value\}.'' \\
    \bottomrule
  \end{tabular}
\end{table}

\textbf{Example injection messages from actual experiments:}
\begin{itemize}
  \item \emph{Goal (format):} ``By the way, please give me the result in CSV format.''
  \item \emph{Constraint (temporal + selection):} ``By the way, I should have mentioned: I'm looking at last month of 2022. Also, 0.50.''
  \item \emph{Input (source):} ``By the way, you can find the data in the shared\_drive/Q3\_budget folder.''
\end{itemize}

\subsubsection{ask\_user Tool Definition}

For natural-ask experiments, the \texttt{ask\_user} tool is provided to models with the following schema (OpenAI function-calling format):

\begin{verbatim}
{
  "name": "ask_user",
  "description": "Ask the user a clarifying question to get
    more information about the task. Use this when the task
    is ambiguous or you need specific details to proceed.",
  "parameters": {
    "properties": {
      "question": {"type": "string",
        "description": "The clarifying question to ask."},
      "context": {"type": "string",
        "description": "Optional additional context.",
        "default": ""}
    },
    "required": ["question"]
  }
}
\end{verbatim}

Additionally, the task prompt is extended with:
\begin{quote}
\small
\texttt{IMPORTANT: Your output will be checked by an auto-grader looking for exact answers. This task may be missing critical information. Use the ask\_user tool to ask the user for any missing details.}
\end{quote}

When the model invokes \texttt{ask\_user}, a simulated user responds with the ground-truth \texttt{removed\_segments} content, matching the injection message that would be used in the forced-injection condition.

\subsubsection{Oracle Budget Calibration}

Injection timing is defined relative to an oracle-calibrated action budget $B_{m,v}$ for each (model $m$, variant $v$) pair. The budget is computed as the mean trajectory length across 3 oracle-condition trials, rounded to the nearest integer:
\[
  B_{m,v} = \text{round}\!\left(\frac{1}{3}\sum_{t=1}^{3} |\tau^{\text{oracle}}_{m,v,t}|\right)
\]
The injection action for timing fraction $f \in \{0.1, 0.3, 0.5, 0.7, 0.9\}$ is:
\[
  a_{\text{inject}} = \max\!\bigl(1,\; \lfloor B_{m,v} \cdot f \rfloor\bigr)
\]

Budget ranges reflect task diversity: MCP-Atlas tasks are short (6--20 actions), TheAgentCompany tasks are moderate (6--49 actions), and SWE-Bench Pro tasks can require extended code navigation and editing (1--121 actions).

\subsubsection{Reproducibility Parameters}
\label{sec:reproducibility}

All models are queried with temperature $= 0.0$ (greedy decoding) and a maximum output length of 4,096 tokens per turn. The three trials per (variant, model, condition) cell use fixed random seeds (0, 1, 2) for any stochastic API parameters. Total API cost for the 6,348 experimental runs was approximately \$4,200 USD across all providers. Task success is determined by the LHAW automated grader~\citep{pu2026lhaw}, which compares agent outputs against ground-truth answers using exact-match or rubric-based criteria depending on the benchmark (MCP-Atlas: tool-output match; TheAgentCompany: checkpoint scoring; SWE-Bench Pro: test-suite pass).

\subsection{Example Trajectory}
\label{sec:example_trajectory}

Table~\ref{fig:example_trace} illustrates the structure of a single forced-injection trial. This MCP-Atlas example (variant \texttt{688ba1...::V\_S1\_delete}, GPT-5.2, injection at 10\%) shows a 7-action trajectory with the injection arriving as a user message after step 1.

\begin{table}[h]
  \caption{Abridged trial trace (MCP-Atlas, injection at 10\%, GPT-5.2). The injection message arrives as a synthetic user turn between steps 1 and 2. All subsequent actions incorporate the clarified information.}
  \label{fig:example_trace}
  \centering
  \small
  \begin{tabular}{clll}
    \toprule
    Step & Phase & Action & Purpose \\
    \midrule
    1 & PRE & \texttt{get\_config} & Environment setup \\
    \multicolumn{4}{c}{\dotfill\ \emph{Injection: ``By the way, I should have mentioned:}} \\
    \multicolumn{4}{c}{\emph{I'm looking at first crime I had registered.''} \dotfill} \\
    2 & POST & \texttt{list\_directory} & Locate data files \\
    3 & POST & \texttt{start\_process} & Launch data tool \\
    4 & POST & \texttt{start\_process} & Execute query \\
    5 & POST & \texttt{search\_pubmed} & Cross-reference \\
    6 & POST & \texttt{search\_pubmed} & Refine search \\
    7 & POST & \texttt{search\_pubmed} & Extract answer \\
    \midrule
    \multicolumn{4}{l}{\textbf{Result:} task\_success = True, total\_actions = 7} \\
    \bottomrule
  \end{tabular}
\end{table}

Each trial JSON records: \texttt{variant\_id}, \texttt{model}, \texttt{condition}, \texttt{injection\_point}, \texttt{actions} (with per-step \texttt{action\_name}, \texttt{parameters}, \texttt{result}, \texttt{is\_pre\_injection}), \texttt{conversation} (full chat history including the injection turn), \texttt{task\_success}, \texttt{total\_actions}, \texttt{pre\_injection\_actions}, \texttt{post\_injection\_actions}, and \texttt{duration\_seconds}. All trial data will be released with the code.

\newpage

\end{document}